\pdfoutput=1
\def\year{2021}\relax
\documentclass[letterpaper]{article} 
\usepackage{aaai21}  
\usepackage{times}  
\usepackage{helvet} 
\usepackage{courier}  
\usepackage[hyphens]{url}  
\usepackage{graphicx} 
\usepackage{natbib}  
\usepackage{caption} 
\usepackage{makecell}
\usepackage{color}
\usepackage{tabularx}
\usepackage{todonotes}
\usepackage{amsmath}
\usepackage{multirow}

\usepackage{xcolor}
\definecolor{ceruleanblue}{rgb}{0.16, 0.32, 0.75}
\usepackage[colorlinks = true,
            linkcolor = blue,
            urlcolor  = blue,
            citecolor = ceruleanblue,
            anchorcolor = blue]{hyperref}

\usepackage[switch]{lineno}  %
\usepackage{lipsum}

\newcommand\blfootnote[1]{%
  \begingroup
  \renewcommand\thefootnote{}\footnote{#1}%
  \addtocounter{footnote}{-1}%
  \endgroup
}

\title{End-to-End QA on COVID-19: Domain Adaptation with Synthetic Training}
\author {

        Revanth Gangi Reddy,\textsuperscript{\rm 1*}
        Bhavani Iyer,\textsuperscript{\rm 2}
        Md Arafat Sultan,\textsuperscript{\rm 2}
        Rong Zhang,\textsuperscript{\rm 2}\\
        Avi Sil,\textsuperscript{\rm 2}
        Vittorio Castelli,\textsuperscript{\rm 2}
        Radu Florian,\textsuperscript{\rm 2}
        Salim Roukos\textsuperscript{\rm 2} \\
}
\affiliations {
    \textsuperscript{\rm 1} University of Illinois, Urbana Champaign \hspace{2em}
    \textsuperscript{\rm 2} IBM Research AI, New York \\
    \texttt{revanth3@illinois.edu, \{bsiyer,zhangr,avi,vittorio,raduf,roukos\}@us.ibm.com, arafat.sultan@ibm.com}
}

\begin{document}
\maketitle

\begin{abstract}
End-to-end question answering (QA) requires both information retrieval (IR) over a large document collection and machine reading comprehension (MRC) on the retrieved passages. Recent work has successfully trained neural IR systems using only supervised question answering (QA) examples from open-domain datasets. However, despite impressive performance on Wikipedia, neural IR lags behind traditional term matching approaches such as BM25 in more specific and specialized target domains such as COVID-19. 
Furthermore, given little or no labeled data, effective adaptation of QA systems can also be challenging in such target domains.
In this work, we explore the application of synthetically generated QA examples to improve performance on closed-domain retrieval and MRC. We combine our neural IR and MRC systems and show significant improvements in end-to-end QA on the CORD-19 collection over a state-of-the-art open-domain QA baseline. 
\end{abstract}

\section{Introduction}
\blfootnote{* Work done during AI Residency at IBM Research.}
In early 2020, the novel coronavirus SARS-CoV-2 was circulating in the U.S. state of New York, unknowingly to its residents, later causing tens of thousands of deaths in the state.
The lack of information available at those early stages of the virus' spread---first about the spread itself, and later about its mitigation including the importance of wearing masks---contributed to more deaths per capita in several northeastern U.S. states than the rest of the country.
This unfortunate occurrence underscores the criticality of early discovery and dissemination of key information in the fight against pandemics such as COVID-19.

End-to-end question answering (QA) systems \cite{lee2019latent,karpukhin2020dense} can be an effective tool for information dissemination in such events.
Given a natural language question, these systems mine related passages from large collections of texts and then extract a short specific answer from the retrieved passages.
Hence they must include both an information retrieval (IR) and a machine reading comprehension (MRC) component. 


Typical approaches to open-domain QA use traditional IR methods---such as TF-IDF matching \cite{chen2017reading} or BM25 term weighting \cite{robertson2009probabilistic}---to retrieve evidence from a corpus. Such IR methods represent text as high-dimensional sparse vectors and rely on inverted indices for efficient search. However, these methods cater almost exclusively to keyword search queries.
Recently, with the advent of large pre-trained language models \cite{radford2019language,devlin2019bert}, using dense semantic representations for IR has emerged as a promising approach. Concurrently, tools like FAISS \cite{johnson2019billion} have been created, which use special in-memory data structures and indexing schemes to provide highly efficient search in a dense vector space. \citeauthor{karpukhin2020dense}~\shortcite{karpukhin2020dense} note that when queried on an index with 21 million passages, FAISS processes 995.0 questions per second, returning top 100 passages per question. In contrast, BM25 processes 23.7 questions per second per CPU thread in a similar setting.

\citeauthor{karpukhin2020dense}~\shortcite{karpukhin2020dense} have shown that neural IR systems trained using human-annotated open-domain MRC data \cite{kwiatkowski2019natural, joshi2017triviaqa} can yield high retrieval
performance on a large collection (Wikipedia, 21 million documents). While such systems have been shown to perform better than BM25 in an open-domain setup, it is unclear how well they would perform in specialized domains where the content and terminology are very specific. Also, such domains often do not have an adequate amount of high-quality annotated data on which neural IR and MRC models can be trained. 
For example, there is very little annotated QA data on COVID-19 although considerable amount of raw text \cite{wang2020cord} is available. 

In this work, we show that automatically generated synthetic training examples from in-domain raw text can yield effective end-to-end QA systems for new target domains, specifically, COVID-19. Our novel synthetic example generator creates training labels for both the IR and the MRC component of the system. Using distant supervision from synthetic question-passage pairs, we fine-tune a pre-trained open-domain neural IR system for the target domain. Synthetic question-passage-answer triples are used to also train a target domain MRC system.
In addition, we adopt existing techniques from the question generation literature such as top-$p$ top-$k$ sampling \cite{sultan-etal-2020-importance} and roundtrip consistency check \cite{alberti2019synthetic} to further improve the quality of generated examples.

In our experiments, we use the scientific articles from the CORD-19 collection \cite{wang2020cord} as the target domain text. 
While recent work \cite{mollercovid, tang2020rapidly, lee2020answering} has released some labeled QA datasets on COVID-19, they are small in size and can only be used to evaluate systems in a zero-shot setting.
This scarcity of human-labeled QA data makes our approach of domain adaptation using synthetic examples especially relevant for this domain. 
However, the techniques proposed in this paper are not specific to COVID-19 and can potentially improve end-to-end QA performance in other domains as well.

Overall, our main contributions in this paper are as follows:
\begin{itemize}
\item We propose a novel approach 
 to generating synthetic training examples for both IR and MRC in target domains where human-annotated data is scarce.
\item We utilize the synthetic examples in a zero-shot domain adaptation setting to separately improve neural IR and MRC performance in the target domain.
\item Finally, combining our improved IR and MRC systems, we show significant improvements in end-to-end QA on multiple COVID-19 datasets over a state-of-the-art open-domain QA system. To our knowledge, our work is the most extensive evaluation of end-to-end QA on a diverse set of QA datasets related to COVID-19.
\end{itemize}

\section{Related Work}

End-to-end QA works in a pipeline where the first important step is to retrieve documents, which potentially contain the answer to an input question, from a large corpus.
Classical sparse vector methods like TF-IDF and BM25 remain strong baselines for IR, and have been used in end-to-end open-domain QA systems \cite{chen2017reading}. More recently, approaches based on dense text representations have emerged as strong alternatives, as they enable modeling of textual similarity at a semantic level. \citeauthor{seo2019real}~\shortcite{seo2019real} propose to combine dense and sparse representations of phrases for real-time question answering. Unsupervised training schemes have also been proposed to latently learn neural retrieval models. \citeauthor{lee2019latent}~\shortcite{lee2019latent} propose an inverse cloze task to train a retriever while \citeauthor{guu2020realm}~\shortcite{guu2020realm} augment masked language model pre-training with a latent knowledge retriever.
\citeauthor{karpukhin2020dense}~\shortcite{karpukhin2020dense} introduce a ``dense passage retriever" (DPR) and show that high performance neural IR models can be trained using labeled MRC data.
Later works \cite{lewis2020retrieval, izacard2020leveraging} have used DPR to obtain competitive open-domain QA performance.
We also adopt DPR as our base IR model and further fine-tune it with target domain examples.

Domain adaptation is an active area of research in natural language processing. Previous work has tackled source-target domain mismatch via instance weighting \cite{jiang2007instance} and training data selection strategies \cite{liu2019reinforced}. \citeauthor{wiese2017neural}~\shortcite{wiese2017neural} have used transfer learning from open-domain datasets. For pre-trained language models, domain adaptation involves further training the model \cite{gururangan-etal-2020-dont} on unlabeled text in the target domain, e.g., scientific articles \cite{beltagy2019scibert} and biomedical text \cite{lee2020biobert, alsentzer2019publicly}.
Here, in addition to fine-tuning on target domain raw text, we generate labeled target domain training data using a synthetic example generator model.

COVID-19 is a low-resource domain with abundant unlabeled text \cite{wang2020cord}. A number of organizations have made their demos\footnote{\href{https://cord19.aws}{Amazon}, \href{https://covid19-research-explorer.appspot.com}{Google}, \href{https://litsearch-site.mybluemix.net}{IBM}, \href{https://cord19.vespa.ai}{Vespa}} available for search over scientific articles on COVID-19. \citeauthor{zhang2020covidex}~\shortcite{zhang2020covidex} build a retrieval system that uses neural ranking models on top of traditional term matching methods. \citeauthor{tang2020rapidly}~\shortcite{tang2020rapidly} use transfer learning from out-of-domain datasets to improve MRC performance. None of  the two systems have been evaluated in an end-to-end QA setting.
We implement a complete QA pipeline for COVID-19 which we evaluate on end-to-end QA and also separately on IR and MRC.

In recent times, pre-trained transformer models have been instrumental in the generation of high quality text across a wide range of applications \cite{radford2019language,lewis-etal-2020-bart,raffel2019exploring}.
In QA, such models as well as older LSTM-based generators have been applied to answer-aware question generation, where the model learns to generate questions from passage-answer pairs \cite{du-cardie-2018-harvesting,unilm,sultan-etal-2020-importance,greater-context}.
We take a similar approach and fine-tune a pre-trained transformer model \cite{lewis-etal-2020-bart} to generate question-answer pairs from passages. We also utilize state-of-the-art techniques such as diversity-promoting sampling \cite{sultan-etal-2020-importance} and roundtrip consistency check \cite{alberti2019synthetic} to further improve performance.


\section{COVID-19 Datasets}

\begin{table*}[ht]
    \centering
    \small
    \begin{tabular}{p{9.3cm}|p{7.7cm}}
    \multicolumn{1}{c|}{\textbf{Passage}}     &  \multicolumn{1}{c}{\textbf{Synthetic Question-Answer pairs}}   \\
    \hline
    ... Since December 2019, when the first patient with a confirmed case of COVID-19 was reported in Wuhan, China, over 1,000,000 patients with confirmed cases have been reported worldwide. It has been reported that the most common symptoms include fever, fatigue, dry cough, anorexia, and dyspnea. Meanwhile, less common symptoms are nasal congestion ...
    & 
    \textbf{Q:} What are the most common symptoms of COVID-19?
         
        \textbf{A:} fever, fatigue, dry cough, anorexia, and dyspnea
        
        ~
        
        \textbf{Q:} How many people have been diagnosed with COVID-19?
        
        \textbf{A:} over 1,000,000\\
        \hline
    ... As with any research, this study is also not without its limitations. First, is the issue of low response rate despite concerted efforts by the research team to contact key informants multiple times. Scholars have argued that such research is often perceived as opportunistic, by the respondents and this perceived lack of trust is likely to have impacted response rates ...
         & \textbf{Q:} What is the main limitation of this study?
         
        \textbf{A:} low response rate
        
        ~
     
        \textbf{Q:} Why was there a low response rate?
        
        \textbf{A:} perceived lack of trust\\
        \hline
    \end{tabular}
    \caption{Synthetic MRC examples generated by our generator from two snippets in the CORD-19 collection.}
    \label{tab:Synth-Ex}
\end{table*}

In this section, we describe our corpus of COVID-19 documents and the
annotated datasets we use for zero-shot evaluation of all
IR and MRC systems.\\

\noindent\textbf{CORD-19}\footnote{\href{https://www.semanticscholar.org/cord19}{https://www.semanticscholar.org/cord19}} \cite{wang2020cord}, or the COVID-19 Open Research Dataset, is a collection of documents taken from the scientific literature on SARS-CoV-2 and other related coronaviruses. The collection is updated on a daily basis; in our experiments, we use 74,059 full text PDFs from its June 22 version. \\



\noindent\textbf{COVID-QA-2019}\footnote{\href{https://github.com/deepset-ai/COVID-QA}{https://github.com/deepset-ai/COVID-QA}} \cite{mollercovid} is a question answering dataset consisting of 2,019 question-article-answer triples. These were created by volunteer biomedical experts from scientific articles related to COVID-19. This dataset differs from traditional
open-domain MRC datasets such as SQuAD \cite{rajpurkar2016squad} in that the examples come from longer contexts (more than 6k words on average vs. SQuAD's 153.2) and answers are also generally longer (13.9 vs. 3.2 words). \\

\noindent\textbf{COVID-QA-147}\footnote{\href{http://covidqa.ai/}{http://covidqa.ai/}} \cite{tang2020rapidly} is a QA dataset obtained from Kaggle's CORD-19 challenge\footnote{\href{https://www.kaggle.com/allen-institute-for-ai/CORD-19-research-challenge}{Kaggle CORD-19}}. Overall, the dataset contains 147 question-article-answer triples with 27 unique questions and 104 unique articles. 
Due to the small number of questions, we only use this dataset for zero-shot evaluation of MRC models on the 147 question-answer pairs.\\

\noindent\textbf{COVID-QA-111}\footnote{\href{https://github.com/dmis-lab/covidAsk}{https://github.com/dmis-lab/covidAsk}} \cite{lee2020answering} contains queries gathered from multiple sources including Kaggle and the FAQ sections of the CDC\footnote{\href{https://www.cdc.gov/coronavirus/2019-ncov/faq.html}{CDC FAQ}} and the WHO\footnote{\href{https://www.who.int/csr/disease/coronavirus_infections/faq_dec12/en/}{WHO FAQ}}. The dataset contains 111 question-answer pairs with 53 interrogative and 58 keyword-style queries. Since questions are not aligned to passages in this dataset, we use it for zero-shot evaluation of only the IR and end-to-end QA systems.

\section{Generating Synthetic Training Examples}

We fine-tune BART~\cite{lewis-etal-2020-bart}---a transformer-based denoising autoencoder with a bidirectional encoder and a causal decoder---to generate synthetic training examples for both IR and MRC.
An MRC example is a triple $(p,q,a)$ comprising a paragraph $p$, a natural language question $q$, and a short answer $a$ (e.g., a named entity) in $p$.
IR training examples consist of only $q$ and $p$.
Previous work has fine-tuned similar transformer models to generate questions from $(p,a)$ pairs, where a separate model extracts a candidate answer $a$ from $p$ prior to question generation \cite{du-cardie-2018-harvesting,alberti2019synthetic,sultan-etal-2020-importance}.

In this work, we train a model to generate both $q$ and $a$ given $p$ in a single pass.
To achieve this, we create training examples for the generator from existing MRC datasets such as SQuAD \cite{rajpurkar2016squad,rajpurkar2018know}.
For each MRC example $(p, q, a)$ where $p$ is the paragraph, $q$ is the question, and $a$ is its human-annotated answer in $p$, we first use a sentence tokenizer to segment $p$ into sentences and locate the sentence $s$ that contains $a$.
Then we create a training example
consisting of $p$ as the source and the ordered sequence $s,a,q$ as the target; special separator tokens separate the three target segments.
A BART autoencoder is then fine-tuned on all such examples to train a generator $g$ with parameters $\theta_g$, which learns to maximize the joint probability $P(s,a,q|p;\theta_g)$.
To speed up generation, we only include the first and the last word of $s$ in the generated version of $s$, which in a vast majority of the cases uniquely disambiguates $s$ in $p$.
In essence, $g$ learns to identify a sentence $s$ in $p$ likely to contain an answer span, extract span $a$ from $s$, and generate the corresponding question $q$, all in a single generative model.

Recently, \citeauthor{sultan-etal-2020-importance}~\shortcite{sultan-etal-2020-importance} have shown that sampling-based question generation leads to better MRC training than greedy or beam search decoding due to increased sample diversity.
We adopt a diversity-promoting top-$p$ top-$k$ sampler for our modified generation task.
Moreover, our design enables the diversification of question-answer pairs as a whole instead of just questions. 
Table~\ref{tab:Synth-Ex} shows some examples of generated QA pairs from two CORD-19 passages.

\section{Information Retrieval}


We adopt a state-of-the-art neural IR model called the Dense Passage Retriever (DPR) \cite{karpukhin2020dense} as our base retrieval model. For COVID-19 domain adaptation, we further fine-tune it on synthetic examples generated from the CORD-19 collection.

\subsection{Dense Passage Retriever (DPR)}
Given a collection of passages, the DPR model creates an index in a continuous space to retrieve passages relevant to an input question.
It uses a Siamese neural network \cite{koch2015siamese} (aka. dual encoder) model with separate dense encoders $E_{Q}(.)$ and $E_{P}(.)$ for the question and passage, respectively. Each encoder is a BERT \cite{devlin2019bert} (base, uncased) model that produces the hidden representation of the [CLS] token as output. The similarity between a question and a passage is the dot product of their encoder outputs:
\begin{equation}\label{eq:sim}
    sim(q,p) = E_Q(q)^TE_P(p)
\end{equation}

\noindent Since Eq.~\ref{eq:sim} is decomposable, representations of all passages in the collection are pre-computed and stored in an index using FAISS \cite{johnson2019billion}. Given an input question \textit{q}, the top \textit{k} passages with representations close to $E_{Q}(q)$ are then retrieved.

\citeauthor{karpukhin2020dense}~\shortcite{karpukhin2020dense} show that training examples for such a dual-encoder model can be obtained from existing MRC datasets. Each training instance $(q_i, p_i^+, p_{i,1}^-, ..., p_{i,n}^-)$ contains a question $q_i$, one positive passage $p_i^+$ and $n$ negative passages $p_{i,j}^-$. The training loss is the negative log-likelihood of the positive passage:
\begin{equation}
    L = -\log\frac{e^{sim(q_i, p_i^+)}}{e^{sim(q_i, p_i^+)} + \sum_{j=1}^n e^{sim(q_i, p_{i,j}^-)}}
\end{equation}

\noindent While negative passages for a given question can be simply sampled from the collection, \citeauthor{karpukhin2020dense}~\shortcite{karpukhin2020dense} show that having a top passage returned by BM25 among the negatives helps improve performance. To make the training process more efficient, the trick of in-batch negatives \cite{yih2011learning, gillick2019learning} is also used. Thus, for each question in a training mini-batch, the following passages are used as negatives: (1)~a passage returned by BM25 that is not labeled positive, (2)~positive passages as well as BM25-retrieved negatives for other questions in the mini-batch. In open-domain QA, DPR outperforms a strong Lucene-BM25 system by 9-19\% top-20 passage retrieval accuracy on a wide range of benchmark datasets.

\subsection{Synthetic Training}
We train our synthetic example generator on existing open-domain MRC data and give it target domain passages as input to produce multiple question-answer pairs per passage.
For IR, we discard the generated answers to first construct positive question-passage pairs.
To create negative examples, for each question, we select a BM25-retrieved passage which does not contain the answer text but has a high lexical overlap with the question.
For each generated question, we create an IR training example by aggregating the question, the positive and the negative passage. During fine-tuning of the DPR model, at each iteration, a set of questions is randomly sampled from the generated dataset. Following \cite{karpukhin2020dense}, we use in-batch negatives while training.  We call this final model the \emph{Adapted DPR} model.


Prior to retrieval using Adapted DPR, we pre-compute the passage representations for the entire retrieval corpus, which for the work presented in this paper was obtained from the CORD-19 collection. The embeddings are indexed using FAISS for efficient run-time retrieval.
Finally, given a question, the same inference procedure as in DPR is followed for retrieval.

\section{Machine Reading Comprehension}

Starting from a pre-trained masked language model (LM), we perform a series of fine-tuning steps to adapt an open-domain MRC system for question answering on CORD-19.
Here we discuss the individual steps in detail as well as the order in which they are applied to construct our final MRC model. 

\subsection{CORD-19 Language Modeling}
\citeauthor{gururangan-etal-2020-dont}~\shortcite{gururangan-etal-2020-dont} have shown that adapting a pre-trained open-domain LM to unlabeled text in a target domain before task-specific fine-tuning can be beneficial for the target task. We begin with a pre-trained RoBERTa-large LM \cite{liu2019roberta} and continue masked LM training \cite{devlin2019bert} on the CORD-19 documents. 
This target domain LM serves as the starting point for the later MRC fine-tuning steps.
 
 \subsection{MRC Model Architecture}
Before giving the details of MRC fine-tuning, we briefly discuss our MRC model architecture.
We use a standard extractive MRC model \cite{devlin2019bert} that extracts a short answer from a passage given a question. The network uses two classification heads on top of a pre-trained RoBERTa LM, which point to the start and end positions of the answer span. For unanswerable examples, the classification heads point to the position of the [CLS] token. Let $start(.)$ and $end(.)$ be the outputs of the start and end classification heads. Then the MRC score of an answer span $(s,e)$, where $s$ is the start and $e$ is the end token, is defined as:
\begin{equation}\label{eq:mrc}
\begin{split}
Sc&ore(s,e) = \\ 
&start(s) + end(e) - start(\mbox{[CLS]}) - end(\mbox{[CLS]})
 \end{split}    
\end{equation}



\subsection{Synthetic Training with Roundtrip Consistency}
To fine-tune the CORD-19 LM for the MRC task, we use both human-annotated data from two open-domain MRC datasets---SQuAD2.0  \cite{rajpurkar2018know} and Natural Questions (NQ) \cite{kwiatkowski2019natural}---and the synthetic question-answer pairs generated by our example generator from CORD-19 passages. 

For the synthetic training examples, we use a roundtrip consistency \cite{alberti2019synthetic} filter to remove noisy examples from the generated data.
It utilizes a pre-trained MRC model to evaluate the quality of automatically generated question-answer pairs. 
Specifically, following \citeauthor{alberti2019bert}~\shortcite{alberti2019bert}, we fine-tune a RoBERTa LM first on SQuAD2.0 and then on NQ.
Given a synthetic question, this MRC model first computes the MRC scores of candidate answer spans (Eq.~\ref{eq:mrc}) in the passage.
We take the highest score over all candidate spans as the \emph{answerability} score of the synthetic question, and filter the example out if this score is lower than a threshold (tuned on our dev fold of the COVID-QA-2019 dataset).
This filter uses the answerability score of the question as a measure of noise in the generated example, since the question generated from the passage is expected to be answerable.

\subsection{Fine-Tuning Sequence}
We adapt the CORD-19 LM to the final MRC task using the following sequence of fine-tuning steps. First, we fine-tune on SQuAD2.0 examples, then on the roundtrip-consistent synthetic examples from the CORD-19 passages, and finally on the NQ examples. We experimented with other sequence orders too but found the above to yield the best performance on our dev fold of the COVID-QA-2019 \cite{mollercovid} dataset. We call this final model the \emph{Adapted MRC} model.

\section{Experimental Setup}

\begin{table*}[ht!]
\centering
\small
\begin{tabular}{l||lll|lll||lll}
\multicolumn{1}{c||}{Model} &  \multicolumn{6}{c||}{Open-COVID-QA-2019} & \multicolumn{3}{c}{COVID-QA-111} \\
\hline

\multicolumn{1}{c||}{} &  \multicolumn{3}{c|}{Dev} & \multicolumn{3}{c||}{Test} & \multicolumn{3}{c}{Test} \\
\multicolumn{1}{c||}{}  & M@20 & M@40 & M@100 & M@20 & M@40 & M@100 & M@20 & M@40 & M@100                         \\ 
                                                           
\hline
BM25 & 22.4 & 24.9 &  29.9&  29.9&  33.4&  39.7&  48.7& 60.4 & 64.9\\
DPR-Multi  &  14.4&  18.4&  22.9&  13.8&  17.5&  21.4& 51.4 &  57.7& 66.7\\
ICT & 16.6 \begin{scriptsize}(1.3)\end{scriptsize} & 21.6 \begin{scriptsize}(1.3)\end{scriptsize} & 25.5 \begin{scriptsize}(0.6)\end{scriptsize} & 18.1 \begin{scriptsize}(0.4)\end{scriptsize} & 23.0 \begin{scriptsize}(0.1)\end{scriptsize} & 29.6 \begin{scriptsize}(0.2)\end{scriptsize} & 52.8 \begin{scriptsize}(0.4)\end{scriptsize} & 59.8 \begin{scriptsize}(1.1)\end{scriptsize} & 67.6 \begin{scriptsize}(2.2)\end{scriptsize} \\
Adapted DPR & 28.0 \begin{scriptsize}(1.8)\end{scriptsize} & 31.8 \begin{scriptsize}(0.8)\end{scriptsize} & 39.0 \begin{scriptsize}(0.5)\end{scriptsize} & 34.8 \begin{scriptsize}(0.3)\end{scriptsize} &  40.4 \begin{scriptsize}(0.2)\end{scriptsize}&  47.2 \begin{scriptsize}(0.2\end{scriptsize})&  58.6 \begin{scriptsize}(0.0)\end{scriptsize}& 64.6 \begin{scriptsize}(1.5)\end{scriptsize} & 74.2 \begin{scriptsize}(1.9)\end{scriptsize}\\
\hline
BM25 + DPR-Multi &  23.4&  27.9&  32.3&  29.5&  33.2&  38.9& 58.6 & 65.8 & 69.4 \\
BM25 + Adapted DPR & \textbf{31.8} \begin{scriptsize}(0.0)\end{scriptsize} & \textbf{36.0} \begin{scriptsize}(0.6)\end{scriptsize} & \textbf{42.6} \begin{scriptsize}(0.6)\end{scriptsize} & \textbf{43.2} \begin{scriptsize}(0.3)\end{scriptsize} & \textbf{48.2} \begin{scriptsize}(0.3)\end{scriptsize} & \textbf{53.7} \begin{scriptsize}(0.3)\end{scriptsize} & \textbf{60.4} \begin{scriptsize}(1.3)\end{scriptsize} & \textbf{68.2} \begin{scriptsize}(0.4)\end{scriptsize} & \textbf{76.9} \begin{scriptsize}(0.8)\end{scriptsize} \\
\hline
\end{tabular}
\caption{Performance of different IR systems on (a) the open version of COVID-QA-2019, and (b) COVID-QA-111.} 
\label{tab:IR}
\end{table*}

\subsection{Datasets}
We use documents from the CORD-19 collection \cite{wang2020cord} to create our retrieval corpus. We split the abstract and the main body of text of each article into passages that (a)~contain no more than 120 words, and (b)~align with sentence boundaries. This leads to an inference-time retrieval corpus of around 3.5 million passages. 

To create the passages from which we generate synthetic training examples, we split the CORD-19 collection into larger chunks of at most 288 wordpieces using the BERT tokenizer, which results in about 1.8 million passages.
This setup provides longer contexts for diverse example generation and also facilitates faster experiments due to a smaller number of passages.

For our MRC experiments, we split the COVID-QA-2019 dataset into dev and test subsets of 203 and 1,816  examples, respectively. Additionally for retrieval and end-to-end QA experiments, we create an open version (Open-COVID-QA-2019 henceforth) wherein duplicate questions are de-duplicated and different answers to the same question are all included in the set of correct answers. This leaves 201 dev and 1,775 test examples in the open version. Finally, we use the COVID-QA-2019 dev set for all hyperparameter tuning experiments, COVID-QA-147 for MRC evaluation, and COVID-QA-111 for evaluating IR and end-to-end QA.

\subsection{Synthetic Example Generation}
To train the synthetic example generator, we use the MRC training examples of SQuAD1.1 \cite{rajpurkar2016squad}. We fine-tune BART for 3 epochs with a learning rate of 3e-5. Using this model, we generate 5 MRC examples from each of the 1.8 million passages in the CORD-19 collection. For top-$p$ top-$k$ sampling, we use $p$=$0.95$ and $k$=$10$. Since it is a generative model, we see cases where the answer text output by the model is not in the input passage. We discard such examples. Overall, the model generates about 7.9 million synthetic examples. We store these as passage-question-answer triples.

\subsection{Neural IR}
As our neural IR baseline, we use the DPR-Multi system---a state-of-the-art neural IR model---from the publicly available implementation\footnote{Code available at \href{https://github.com/facebookresearch/DPR}{https://github.com/facebookresearch/DPR}} provided by \citeauthor{karpukhin2020dense}~\shortcite{karpukhin2020dense}. This system comes pre-trained on the open versions of multiple MRC datasets: Natural Questions~\cite{kwiatkowski2019natural}, WebQuestions~\cite{berant2013semantic}, CuratedTrec~\cite{baudivs2015modeling} and TriviaQA~\cite{joshi2017triviaqa}. 

We fine-tune the DPR-Multi system for 6 epochs using the synthetic examples with a learning rate of 1e-5 and a batch size of 128. The resulting model is our Adapted DPR model.

We also use a second neural IR baseline based on the Inverse Cloze Task (ICT) method proposed in \cite{lee2019latent}. ICT is an unsupervised training procedure wherein a sentence is randomly masked out from the passage with a probability $p$ and used as the query to create a query-passage synthetic training pair. We adopt ICT as an alternative  approach to generating synthetic training examples. We set $p$=$0.9$, which \citeauthor{lee2019latent}~\shortcite{lee2019latent} have shown to work best, and use the 288 wordpiece passages from the CORD-19 collection to create 1.8 million training examples. We train for 6 epochs using these ICT examples with a learning rate of 1e-5 and a batch size of 128. We then follow \citeauthor{lee2019latent}~\shortcite{lee2019latent} to do a final round of fine-tuning wherein only the question encoder is trained for 10 epochs using questions from the open version of NQ. Since the above technique does not require any in-domain labeled data, we use it as a baseline domain adaptation approach. We call this model the \textit{ICT} model.


\begin{table*}[t]
    \centering
    \small
    \begin{tabular}{p{3.7cm}|p{6.4cm}|p{6.4cm}}
    \multicolumn{1}{c|}{\textbf{Example}}     &  \multicolumn{1}{c|}{\textbf{Adapted DPR}} &  \multicolumn{1}{c}{\textbf{BM25}} \\
    \hline
    \textbf{Q:} What was the fatality rate for SARS-CoV?
    
    \textbf{A:} 10\%
    & The case fatality rate (CFR) of COVID-19 was 2.3\% (44/1023), much lower than that of SARS (\textbf{10\%}) and MERS (36\%) (de Wit et al. 2016; Wu and McGoogan 2020). Suspected COVID-19 patients (with symptoms) could be diagnosed ...
    & 
    The analysis estimated that the case-fatality rate of COVID-19 in Europe would range between 4\% and 4.5\%. The case-fatality rate of SARS-COV, which was a similar outbreak, was \textbf{10\%}, while the case-fatality rate of MERS-CoV was over 35\% ...\\
    \hline
    \textbf{Q:} What is the molecular structure of the Human metapneumovirus (HMPV)?
    
    \textbf{A:} single-stranded RNA virus
    &
    Human bocavirus: hMPV is a paramyxovirus first discovered by van den Hoogen and colleagues15 in 2001. Similar to RSV, hMPV is a \textbf{single-stranded RNA virus} belonging to the Pneumoviridae subfamily, and causes many of the same symptoms ...
    &
    ... in specimens from 1976 to 2001. Collectively, these studies show that HMPV has been circulating undetected for many decades. Genome organization and structure: HMPV is a negative-sense, non-segmented, \textbf{single-stranded RNA virus}.
    \\
    \hline
    \end{tabular}
    \caption{Examples where Adapted DPR and BM25 both retrieve passages that are not returned by the other system (in the top 100 results).}
    \label{tab:Retrieved-Ex}
\end{table*}

\subsection{Machine Reading Comprehension}
The baseline MRC system fine-tunes a pre-trained RoBERTa-large model for 3 epochs on SQuAD2.0 and then for 1 epoch on Natural Questions (NQ) training examples. It achieves a short answer EM of 59.4 on the NQ dev set, which is competitive with numbers reported in \cite{liu2020rikinet}. We use the Transformers library \cite{Wolf2019HuggingFacesTS} for all our MRC experiments. 

For masked LM fine-tuning of the pre-trained RoBERTa-large LM on the CORD-19 collection, we use approximately 1.5GB of text containing 225 million tokens.
We train for 8 epochs with a learning rate of 1.5e-4 using the Fairseq toolkit \cite{ott2019fairseq}. For the downstream fine-tuning of this LM to the MRC task, we train for 3 epochs on SQuAD2.0, 1 epoch each on the filtered synthetic MRC examples and the NQ dataset. During roundtrip consistency filtering, we use a high answerability score threshold of $t$=$7.0$, and are left with around 380k synthetic MRC examples after filtering. 

\subsection{Metrics}

 We evaluate the IR models using the recall-based Match@20, Match@40 and Match@100 metrics, similar to \cite{karpukhin2020dense}. These metrics measure the top-$k$ retrieval accuracy, which is the fraction of questions for which the top $k$ retrieved passages contain a span that answers the question. For the MRC models, we use the standard Exact Match (EM) and F1 score for evaluation. Finally, we evaluate the end-to-end QA systems on Top-1 F1 and Top-5 F1.

\section{Results and Discussion}

In this section, we first report results separately for our IR and MRC systems. Then we evaluate the end-to-end QA system which combines the IR and the MRC component and uses the entire CORD-19 collection to find an answer given an input question.
Reported numbers for all our trained models are averages over three seeds.


\subsection{Information Retrieval} 

We evaluate our proposed system, Adapted DPR, against a number of traditional term matching and neural IR baselines. Specifically, we use BM25\footnote{\href{https://lucene.apache.org}{Lucene} Implementation. BM25 parameters $b=0.75$ (document length normalization) and $k_1=1.2$ (term frequency scaling) worked best.} as the term matching baseline, the DPR-Multi system as the zero-shot open-domain baseline and ICT as a domain adaptation baseline. 
Further, we also evaluate models that combine term matching and neural approaches.
We take the top 2,000 passages retrieved by BM25 and neural models separately, and score each passage using a convex combination of its BM25 and neural IR scores after normalization (weight is tuned on the Open-COVID-QA-2019 dev set). We create two such combined systems: BM25+DPR-Multi and BM25+Adapted DPR.

\subsubsection{Results}
Table \ref{tab:IR} shows the performance of different IR systems on the Open-COVID-QA-2019 and COVID-QA-111 datasets.
BM25 shows strong performance on both datasets, demonstrating the robustness of such term matching methods. While the neural DPR-multi system is competitive with BM25 on COVID-QA-111, it is considerably behind on the larger Open-COVID-QA-2019 dataset. The ICT model improves over DPR-multi, showing that domain adaption using such unsupervised techniques can be beneficial. 

Our Adapted DPR system achieves the best single system results on both datasets, demonstrating the effectiveness of using our synthetic example generation for domain adaptation. 
On the Open-COVID-QA-2019 test set, our model improves over the baseline DPR-Multi system by more than 100\%.
Finally, we see that a combination of BM25 and the neural approaches can give considerable performance improvements. 
Combining DPR-Multi with BM25 does not lead to any gains on Open-COVID-QA-2019 likely due to the fact that DPR-Multi performs poorly on this dataset.
However, we see large gains from combining BM25 and our Adapted DPR system, as both perform well individually on the two datasets. Our final BM25+Adapted DPR system is better than the next best baseline by about 14 points across all metrics on the test set of Open-COVID-QA-2019 and up to 7 points on Match@100 on COVID-QA-111.

\subsubsection{Analysis}
On a closer look at the passages retrieved individually by BM25 and Adapted DPR, we observe that the two sets of retrieved passages are very different. For the Open-COVID-QA-2019 dataset, only 5 passages are common on average among the top 100 passages retrieved separately by the two systems. This difference is also visible in the relevant passages (passages that contain an answer to the question) that are returned by the two systems. We observe many cases where the two systems retrieve mutually exclusive relevant passages in the top 100 retrieved results. Table \ref{tab:Retrieved-Ex} shows two such examples. This diversity in retrieval results demonstrates the complementary nature of the two systems and also explains why their combination leads to improved IR performance.

\begin{table}[h]
\centering
\begin{tabular}{llll}

\multicolumn{1}{c}{Model} & M@20 & M@40 & M@100 \\ 
\hline
NQ-style SynQ & 20.4 & 23.9 &   27.9 \\
Squad-style SynQ & 28.0 & 31.8 &  39.0  \\
\hline
\end{tabular}
\caption{Retrieval performance on the dev fold of Open-COVID-QA-2019.} 
\label{tab:IR_Qtype}
\end{table}

\begin{table*}[!]
\centering
\begin{tabular}{l||ll|ll||ll}
\multicolumn{1}{c||}{Model} & \multicolumn{4}{c||}{Open-COVID-QA-2019} & \multicolumn{2}{c}{COVID-QA-111} \\
\hline
\multicolumn{1}{c||}{} & \multicolumn{2}{c|}{Dev} & \multicolumn{2}{c||}{Test} & \multicolumn{2}{c}{Test} \\
\multicolumn{1}{c||}{} & Top-1 & Top-5 & Top-1 & Top-5 & Top-1 & Top-5 \\
\hline
BM25 $\rightarrow$ Baseline MRC & 21.7 & 31.8 & 27.1 & 38.7  & 24.1 & 39.3\\
(BM25 + DPR-Multi) $\rightarrow$ Baseline MRC & 21.4 & 30.9 & 25.2 & 37.2 & 24.4 & 43.2 \\
(BM25 + Adapted DPR) $\rightarrow$ Baseline MRC & 24.2 \begin{scriptsize}(0.9)\end{scriptsize}  & 35.6 \begin{scriptsize}(0.3)\end{scriptsize}  & 29.5 \begin{scriptsize}(0.1)\end{scriptsize} & 44.2 \begin{scriptsize}(0.2)\end{scriptsize}  & 25.0 \begin{scriptsize}(0.2)\end{scriptsize} & 45.9 \begin{scriptsize}(0.8)\end{scriptsize}  \\
(BM25 + Adapted DPR) $\rightarrow$ Adapted MRC & \textbf{27.2} \begin{scriptsize}(0.9)\end{scriptsize} & \textbf{37.2} \begin{scriptsize}(0.2)\end{scriptsize} &  \textbf{30.4} \begin{scriptsize}(0.3)\end{scriptsize} & \textbf{44.9} \begin{scriptsize}(0.1)\end{scriptsize} & \textbf{26.5} \begin{scriptsize}(0.5)\end{scriptsize} & \textbf{47.8} \begin{scriptsize}(0.8)\end{scriptsize} \\
\hline
\end{tabular}
\caption{End-to-end question answering F1 scores on the open version of COVID-QA-2019 and COVID-QA-111.} 
\label{tab:Open_QA}
\end{table*}

To further investigate synthetic example generation, besides SQuAD, we also train the generator on a second MRC dataset, namely, the Natural Questions (NQ) dataset \cite{kwiatkowski2019natural}. NQ contains information seeking questions from real-life users of Google search whereas SQuAD contains well-formed questions created by annotators after looking at the passage. Thus these two datasets contain distinct question styles; we explore which one yields a better synthetic example generator for our application. Table \ref{tab:IR_Qtype} compares the performance of the NQ-style synthetic examples with the SQuAD-style examples while adapting the DPR-Multi model. We can see from the results that using questions from a SQuAD-trained synthetic generator is considerably better.


\subsection{Machine Reading Comprehension}
Table \ref{tab:QA} shows results on the test sets of different MRC datasets. Input to the models is a question and an annotated document that contains an answer. The Adapted MRC model incorporates both language modeling on the CORD-19 collection and synthetic MRC training. Over our state-of-the-art open-domain MRC baseline, we see 2 and 3.7 F1 improvements on test sets of COVID-QA-2019 and COVID-QA-147, respectively.

\begin{table}[h]
\centering
\small
\begin{tabular}{lllll}
\multicolumn{1}{c}{Model} & \multicolumn{2}{c}{COVID-QA-2019} & \multicolumn{2}{c}{COVID-QA-147} \\ 
\hline
\multicolumn{1}{c}{} & EM & F1 &  EM & F1                          \\ 
                                                           
\hline
Baseline MRC  & 34.7 &  62.7 & 8.8 & 31.0\\
Adapted MRC   & \textbf{37.2} \begin{scriptsize}(0.4)\end{scriptsize} &  \textbf{64.7} \begin{scriptsize}(0.1)\end{scriptsize} & \textbf{11.3} \begin{scriptsize}(0.6)\end{scriptsize} & \textbf{34.7} \begin{scriptsize}(1.1)\end{scriptsize}\\
\hline
\end{tabular}
\caption{MRC performance on the test folds of two datasets.} 
\label{tab:QA}
\end{table}

To further demonstrate the improvements from language modeling and using synthetic examples, we present in Table \ref{tab:QA_dev} the results on the COVID-QA-2019 dev set from incrementally applying the two domain adaptation strategies. We see that both strategies yield performance gains. However, it can be seen that using the synthetic MRC examples from our  generator contributes more, with 3.1  EM and 2.6 F1 increase vs 1.5 EM and 0.8 F1 improvement from language modeling.

\begin{table}[h]
\centering
\begin{tabular}{lcc}
\multicolumn{1}{c}{Model} & EM & F1 \\
\hline                       
                                                           
\hline
Baseline MRC & 34.0 & 59.4   \\
+ Language Modeling on CORD-19  & 35.5 & 60.2   \\
+ Adding SynQ during MRC training & 38.6 & 62.8    \\
\hline
\end{tabular}
\caption{Machine reading comprehension performance on the dev split of COVID-QA-2019.} 
\label{tab:QA_dev}
\end{table}

\subsection{End-to-End Question Answering}

Finally, we combine different IR and MRC systems to create end-to-end QA systems. We measure the improvements from our domain adaptation strategy in this setting, where only the question is given as input for QA over the entire corpus. 

In the retrieval phase, we take the top $K$ passages ($K$ tuned on dev) from the IR system. Each passage is then passed to the MRC model to get the top answer and its MRC score. Finally, we normalize the IR and MRC scores and combine via a convex combination (IR weight = $0.7$, tuned on dev). We observe that using $K$=$100$ works best when IR is BM25 only and $K$=$40$ works best for BM25 + Neural IR systems.
Table \ref{tab:Open_QA} shows the end-to-end F1 performance of the combination of IR and MRC systems. 
We see that both having a better retriever (BM25+Adapted DPR) and a better MRC (Adpated MRC) model contribute to improvements in end-to-end QA performance.

To verify the statistical significance of our end-to-end QA results, we perform a paired $t$-test \cite{hsu2005paired} on the Top-5 F1 scores for both datasets. Our final end-to-end QA system is significantly better than the baseline system at $p<0.01$.

\section{Conclusion}

We present an approach for zero-shot adaptation of an open-domain end-to-end question answering system to a target domain, in this case COVID-19. We propose a novel example generation model that can produce synthetic training examples for both information retrieval and machine reading comprehension. Importantly, our generation model, trained using open-domain supervised QA data, is used to generate synthetic question-answer pairs in the target domain.
By running extensive evaluation experiments, we show that our end-to-end QA model as well as its individual IR and MRC components benefit from the synthetic examples.

Low-resource target domains can present significant challenges for natural language processing systems.
Our work shows that synthetic generation can be an effective domain adaptation approach for QA.
Future work will explore semi-supervised and active learning approaches to determine how a small amount supervision can further improve results.

\bibliographystyle{aaai21}
\bibliography{aaai2021}

\begin{thebibliography}{45}
\providecommand{\natexlab}[1]{#1}
\providecommand{\url}[1]{\texttt{#1}}
\providecommand{\urlprefix}{URL }
\expandafter\ifx\csname urlstyle\endcsname\relax
  \providecommand{\doi}[1]{doi:\discretionary{}{}{}#1}\else
  \providecommand{\doi}{doi:\discretionary{}{}{}\begingroup
  \urlstyle{rm}\Url}\fi

\bibitem[{Alberti et~al.(2019)Alberti, Andor, Pitler, Devlin, and
  Collins}]{alberti2019synthetic}
Alberti, C.; Andor, D.; Pitler, E.; Devlin, J.; and Collins, M. 2019.
\newblock Synthetic QA corpora generation with roundtrip consistency.
\newblock \emph{arXiv preprint arXiv:1906.05416} .

\bibitem[{Alberti, Lee, and Collins(2019)}]{alberti2019bert}
Alberti, C.; Lee, K.; and Collins, M. 2019.
\newblock A bert baseline for the natural questions.
\newblock \emph{arXiv preprint arXiv:1901.08634} .

\bibitem[{Alsentzer et~al.(2019)Alsentzer, Murphy, Boag, Weng, Jindi, Naumann,
  and McDermott}]{alsentzer2019publicly}
Alsentzer, E.; Murphy, J.; Boag, W.; Weng, W.-H.; Jindi, D.; Naumann, T.; and
  McDermott, M. 2019.
\newblock Publicly Available Clinical BERT Embeddings.
\newblock In \emph{Proceedings of the 2nd Clinical Natural Language Processing
  Workshop}, 72--78.

\bibitem[{Baudi{\v{s}} and {\v{S}}ediv{\`y}(2015)}]{baudivs2015modeling}
Baudi{\v{s}}, P.; and {\v{S}}ediv{\`y}, J. 2015.
\newblock Modeling of the question answering task in the yodaqa system.
\newblock In \emph{International Conference of the Cross-Language Evaluation
  Forum for European Languages}, 222--228. Springer.

\bibitem[{Beltagy, Lo, and Cohan(2019)}]{beltagy2019scibert}
Beltagy, I.; Lo, K.; and Cohan, A. 2019.
\newblock SciBERT: A Pretrained Language Model for Scientific Text.
\newblock In \emph{Proceedings of the 2019 Conference on Empirical Methods in
  Natural Language Processing and the 9th International Joint Conference on
  Natural Language Processing (EMNLP-IJCNLP)}, 3606--3611.

\bibitem[{Berant et~al.(2013)Berant, Chou, Frostig, and
  Liang}]{berant2013semantic}
Berant, J.; Chou, A.; Frostig, R.; and Liang, P. 2013.
\newblock Semantic parsing on freebase from question-answer pairs.
\newblock In \emph{Proceedings of the 2013 conference on empirical methods in
  natural language processing}, 1533--1544.

\bibitem[{Chen et~al.(2017)Chen, Fisch, Weston, and Bordes}]{chen2017reading}
Chen, D.; Fisch, A.; Weston, J.; and Bordes, A. 2017.
\newblock Reading {Wikipedia} to Answer Open-Domain Questions.
\newblock In \emph{Association for Computational Linguistics (ACL)}.

\bibitem[{Devlin et~al.(2019)Devlin, Chang, Lee, and
  Toutanova}]{devlin2019bert}
Devlin, J.; Chang, M.-W.; Lee, K.; and Toutanova, K. 2019.
\newblock BERT: Pre-training of Deep Bidirectional Transformers for Language
  Understanding.
\newblock In \emph{Proceedings of the 2019 Conference of the North American
  Chapter of the Association for Computational Linguistics: Human Language
  Technologies, Volume 1 (Long and Short Papers)}, 4171--4186.

\bibitem[{Dong et~al.(2019)Dong, Yang, Wang, Wei, Liu, Wang, Gao, Zhou, and
  Hon}]{unilm}
Dong, L.; Yang, N.; Wang, W.; Wei, F.; Liu, X.; Wang, Y.; Gao, J.; Zhou, M.;
  and Hon, H.-W. 2019.
\newblock Unified Language Model Pre-training for Natural Language
  Understanding and Generation.
\newblock In \emph{Proceedings of NeurIPS}.

\bibitem[{Du and Cardie(2018)}]{du-cardie-2018-harvesting}
Du, X.; and Cardie, C. 2018.
\newblock Harvesting Paragraph-level Question-Answer Pairs from {W}ikipedia.
\newblock In \emph{Proceedings of the 56th Annual Meeting of the Association
  for Computational Linguistics (Volume 1: Long Papers)}, 1907--1917.
  Association for Computational Linguistics.

\bibitem[{Gillick et~al.(2019)Gillick, Kulkarni, Lansing, Presta, Baldridge,
  Ie, and Garcia-Olano}]{gillick2019learning}
Gillick, D.; Kulkarni, S.; Lansing, L.; Presta, A.; Baldridge, J.; Ie, E.; and
  Garcia-Olano, D. 2019.
\newblock Learning Dense Representations for Entity Retrieval.
\newblock In \emph{Proceedings of the 23rd Conference on Computational Natural
  Language Learning (CoNLL)}, 528--537.

\bibitem[{Gururangan et~al.(2020)Gururangan, Marasovi{\'c}, Swayamdipta, Lo,
  Beltagy, Downey, and Smith}]{gururangan-etal-2020-dont}
Gururangan, S.; Marasovi{\'c}, A.; Swayamdipta, S.; Lo, K.; Beltagy, I.;
  Downey, D.; and Smith, N.~A. 2020.
\newblock Don{'}t Stop Pretraining: Adapt Language Models to Domains and Tasks.
\newblock In \emph{Proceedings of the 58th Annual Meeting of the Association
  for Computational Linguistics}, 8342--8360.

\bibitem[{Guu et~al.(2020)Guu, Lee, Tung, Pasupat, and Chang}]{guu2020realm}
Guu, K.; Lee, K.; Tung, Z.; Pasupat, P.; and Chang, M.-W. 2020.
\newblock Realm: Retrieval-augmented language model pre-training.
\newblock \emph{arXiv preprint arXiv:2002.08909} .

\bibitem[{Hsu and Lachenbruch(2005)}]{hsu2005paired}
Hsu, H.; and Lachenbruch, P.~A. 2005.
\newblock Paired t test.
\newblock \emph{Encyclopedia of Biostatistics} 6.

\bibitem[{Izacard and Grave(2020)}]{izacard2020leveraging}
Izacard, G.; and Grave, E. 2020.
\newblock Leveraging Passage Retrieval with Generative Models for Open Domain
  Question Answering.
\newblock \emph{arXiv preprint arXiv:2007.01282} .

\bibitem[{Jiang and Zhai(2007)}]{jiang2007instance}
Jiang, J.; and Zhai, C. 2007.
\newblock Instance weighting for domain adaptation in NLP.
\newblock In \emph{Proceedings of the 45th annual meeting of the association of
  computational linguistics}, 264--271.

\bibitem[{Johnson, Douze, and J{\'e}gou(2019)}]{johnson2019billion}
Johnson, J.; Douze, M.; and J{\'e}gou, H. 2019.
\newblock Billion-scale similarity search with GPUs.
\newblock \emph{IEEE Transactions on Big Data} .

\bibitem[{Joshi et~al.(2017)Joshi, Choi, Weld, and
  Zettlemoyer}]{joshi2017triviaqa}
Joshi, M.; Choi, E.; Weld, D.~S.; and Zettlemoyer, L. 2017.
\newblock TriviaQA: A Large Scale Distantly Supervised Challenge Dataset for
  Reading Comprehension.
\newblock In \emph{Proceedings of the 55th Annual Meeting of the Association
  for Computational Linguistics (Volume 1: Long Papers)}, 1601--1611.

\bibitem[{Karpukhin et~al.(2020)Karpukhin, O{\u{g}}uz, Min, Wu, Edunov, Chen,
  and Yih}]{karpukhin2020dense}
Karpukhin, V.; O{\u{g}}uz, B.; Min, S.; Wu, L.; Edunov, S.; Chen, D.; and Yih,
  W.-t. 2020.
\newblock Dense Passage Retrieval for Open-Domain Question Answering.
\newblock \emph{arXiv preprint arXiv:2004.04906} .

\bibitem[{Koch, Zemel, and Salakhutdinov(2015)}]{koch2015siamese}
Koch, G.; Zemel, R.; and Salakhutdinov, R. 2015.
\newblock Siamese neural networks for one-shot image recognition.
\newblock In \emph{ICML deep learning workshop}, volume~2. Lille.

\bibitem[{Kwiatkowski et~al.(2019)Kwiatkowski, Palomaki, Redfield, Collins,
  Parikh, Alberti, Epstein, Polosukhin, Devlin, Lee
  et~al.}]{kwiatkowski2019natural}
Kwiatkowski, T.; Palomaki, J.; Redfield, O.; Collins, M.; Parikh, A.; Alberti,
  C.; Epstein, D.; Polosukhin, I.; Devlin, J.; Lee, K.; et~al. 2019.
\newblock Natural questions: a benchmark for question answering research.
\newblock \emph{Transactions of the Association for Computational Linguistics}
  7: 453--466.

\bibitem[{Lee et~al.(2020{\natexlab{a}})Lee, Yi, Jeong, Sung, Yoon, Choi, Ko,
  and Kang}]{lee2020answering}
Lee, J.; Yi, S.~S.; Jeong, M.; Sung, M.; Yoon, W.; Choi, Y.; Ko, M.; and Kang,
  J. 2020{\natexlab{a}}.
\newblock Answering Questions on COVID-19 in Real-Time.
\newblock \emph{arXiv preprint arXiv:2006.15830} .

\bibitem[{Lee et~al.(2020{\natexlab{b}})Lee, Yoon, Kim, Kim, Kim, So, and
  Kang}]{lee2020biobert}
Lee, J.; Yoon, W.; Kim, S.; Kim, D.; Kim, S.; So, C.; and Kang, J.
  2020{\natexlab{b}}.
\newblock BioBERT: a pre-trained biomedical language representation model for
  biomedical text mining.
\newblock \emph{Bioinformatics (Oxford, England)} 36(4): 1234.

\bibitem[{Lee, Chang, and Toutanova(2019)}]{lee2019latent}
Lee, K.; Chang, M.-W.; and Toutanova, K. 2019.
\newblock Latent Retrieval for Weakly Supervised Open Domain Question
  Answering.
\newblock In \emph{Proceedings of the 57th Annual Meeting of the Association
  for Computational Linguistics}, 6086--6096.

\bibitem[{Lewis et~al.(2020{\natexlab{a}})Lewis, Liu, Goyal, Ghazvininejad,
  Mohamed, Levy, Stoyanov, and Zettlemoyer}]{lewis-etal-2020-bart}
Lewis, M.; Liu, Y.; Goyal, N.; Ghazvininejad, M.; Mohamed, A.; Levy, O.;
  Stoyanov, V.; and Zettlemoyer, L. 2020{\natexlab{a}}.
\newblock {BART}: Denoising Sequence-to-Sequence Pre-training for Natural
  Language Generation, Translation, and Comprehension.
\newblock In \emph{Proceedings of the 58th Annual Meeting of the Association
  for Computational Linguistics}, 7871--7880. Association for Computational
  Linguistics.

\bibitem[{Lewis et~al.(2020{\natexlab{b}})Lewis, Perez, Piktus, Petroni,
  Karpukhin, Goyal, K{\"u}ttler, Lewis, Yih, Rockt{\"a}schel
  et~al.}]{lewis2020retrieval}
Lewis, P.; Perez, E.; Piktus, A.; Petroni, F.; Karpukhin, V.; Goyal, N.;
  K{\"u}ttler, H.; Lewis, M.; Yih, W.-t.; Rockt{\"a}schel, T.; et~al.
  2020{\natexlab{b}}.
\newblock Retrieval-augmented generation for knowledge-intensive nlp tasks.
\newblock \emph{arXiv preprint arXiv:2005.11401} .

\bibitem[{Liu et~al.(2020)Liu, Gong, Fu, Yan, Chen, Jiang, Lv, and
  Duan}]{liu2020rikinet}
Liu, D.; Gong, Y.; Fu, J.; Yan, Y.; Chen, J.; Jiang, D.; Lv, J.; and Duan, N.
  2020.
\newblock RikiNet: Reading Wikipedia Pages for Natural Question Answering.
\newblock \emph{arXiv preprint arXiv:2004.14560} .

\bibitem[{Liu et~al.(2019{\natexlab{a}})Liu, Song, Zou, and
  Zhang}]{liu2019reinforced}
Liu, M.; Song, Y.; Zou, H.; and Zhang, T. 2019{\natexlab{a}}.
\newblock Reinforced training data selection for domain adaptation.
\newblock In \emph{Proceedings of the 57th Annual Meeting of the Association
  for Computational Linguistics}, 1957--1968.

\bibitem[{Liu et~al.(2019{\natexlab{b}})Liu, Ott, Goyal, Du, Joshi, Chen, Levy,
  Lewis, Zettlemoyer, and Stoyanov}]{liu2019roberta}
Liu, Y.; Ott, M.; Goyal, N.; Du, J.; Joshi, M.; Chen, D.; Levy, O.; Lewis, M.;
  Zettlemoyer, L.; and Stoyanov, V. 2019{\natexlab{b}}.
\newblock Roberta: A robustly optimized bert pretraining approach.
\newblock \emph{arXiv preprint arXiv:1907.11692} .

\bibitem[{M{\"o}ller et~al.(2020)M{\"o}ller, Reina, Jayakumar, and
  Livermore}]{mollercovid}
M{\"o}ller, T.; Reina, G.~A.; Jayakumar, R.; and Livermore, L. 2020.
\newblock COVID-QA: A Question Answering Dataset for COVID-19 .

\bibitem[{Ott et~al.(2019)Ott, Edunov, Baevski, Fan, Gross, Ng, Grangier, and
  Auli}]{ott2019fairseq}
Ott, M.; Edunov, S.; Baevski, A.; Fan, A.; Gross, S.; Ng, N.; Grangier, D.; and
  Auli, M. 2019.
\newblock fairseq: A Fast, Extensible Toolkit for Sequence Modeling.
\newblock In \emph{Proceedings of NAACL-HLT 2019: Demonstrations}.

\bibitem[{Radford et~al.(2019)Radford, Wu, Child, Luan, Amodei, and
  Sutskever}]{radford2019language}
Radford, A.; Wu, J.; Child, R.; Luan, D.; Amodei, D.; and Sutskever, I. 2019.
\newblock Language Models are Unsupervised Multitask Learners .

\bibitem[{Raffel et~al.(2019)Raffel, Shazeer, Roberts, Lee, Narang, Matena,
  Zhou, Li, and Liu}]{raffel2019exploring}
Raffel, C.; Shazeer, N.; Roberts, A.; Lee, K.; Narang, S.; Matena, M.; Zhou,
  Y.; Li, W.; and Liu, P.~J. 2019.
\newblock Exploring the limits of transfer learning with a unified text-to-text
  transformer.
\newblock \emph{arXiv preprint arXiv:1910.10683} .

\bibitem[{Rajpurkar, Jia, and Liang(2018)}]{rajpurkar2018know}
Rajpurkar, P.; Jia, R.; and Liang, P. 2018.
\newblock Know What You Don’t Know: Unanswerable Questions for SQuAD.
\newblock In \emph{Proceedings of the 56th Annual Meeting of the Association
  for Computational Linguistics (Volume 2: Short Papers)}, 784--789.

\bibitem[{Rajpurkar et~al.(2016)Rajpurkar, Zhang, Lopyrev, and
  Liang}]{rajpurkar2016squad}
Rajpurkar, P.; Zhang, J.; Lopyrev, K.; and Liang, P. 2016.
\newblock SQuAD: 100,000+ Questions for Machine Comprehension of Text.
\newblock In \emph{Proceedings of the 2016 Conference on Empirical Methods in
  Natural Language Processing}, 2383--2392.

\bibitem[{Robertson and Zaragoza(2009)}]{robertson2009probabilistic}
Robertson, S.; and Zaragoza, H. 2009.
\newblock The Probabilistic Relevance Framework: BM25 and Beyond.
\newblock \emph{Foundations and Trends in Information Retrieval} 3(4):
  333--389.

\bibitem[{Seo et~al.(2019)Seo, Lee, Kwiatkowski, Parikh, Farhadi, and
  Hajishirzi}]{seo2019real}
Seo, M.; Lee, J.; Kwiatkowski, T.; Parikh, A.; Farhadi, A.; and Hajishirzi, H.
  2019.
\newblock Real-Time Open-Domain Question Answering with Dense-Sparse Phrase
  Index.
\newblock In \emph{Proceedings of the 57th Annual Meeting of the Association
  for Computational Linguistics}, 4430--4441.

\bibitem[{Sultan et~al.(2020)Sultan, Chandel, Fernandez~Astudillo, and
  Castelli}]{sultan-etal-2020-importance}
Sultan, M.~A.; Chandel, S.; Fernandez~Astudillo, R.; and Castelli, V. 2020.
\newblock On the Importance of Diversity in Question Generation for {QA}.
\newblock In \emph{Proceedings of the 58th Annual Meeting of the Association
  for Computational Linguistics}, 5651--5656. Association for Computational
  Linguistics.

\bibitem[{Tang et~al.(2020)Tang, Nogueira, Zhang, Gupta, Cam, Cho, and
  Lin}]{tang2020rapidly}
Tang, R.; Nogueira, R.; Zhang, E.; Gupta, N.; Cam, P.; Cho, K.; and Lin, J.
  2020.
\newblock Rapidly Bootstrapping a Question Answering Dataset for COVID-19.
\newblock \emph{arXiv preprint arXiv:2004.11339} .

\bibitem[{Tuan, Shah, and Barzilay(2020)}]{greater-context}
Tuan, L.~A.; Shah, D.~J.; and Barzilay, R. 2020.
\newblock Capturing Greater Context for Question Generation.
\newblock In \emph{Proceedings of AAAI}.

\bibitem[{Wang et~al.(2020)Wang, Lo, Chandrasekhar, Reas, Yang, Eide, Funk,
  Kinney, Liu, Merrill et~al.}]{wang2020cord}
Wang, L.~L.; Lo, K.; Chandrasekhar, Y.; Reas, R.; Yang, J.; Eide, D.; Funk, K.;
  Kinney, R.; Liu, Z.; Merrill, W.; et~al. 2020.
\newblock CORD-19: The Covid-19 Open Research Dataset.
\newblock \emph{ArXiv} .

\bibitem[{Wiese, Weissenborn, and Neves(2017)}]{wiese2017neural}
Wiese, G.; Weissenborn, D.; and Neves, M. 2017.
\newblock Neural Domain Adaptation for Biomedical Question Answering.
\newblock In \emph{Proceedings of the 21st Conference on Computational Natural
  Language Learning (CoNLL 2017)}, 281--289.

\bibitem[{Wolf et~al.(2019)Wolf, Debut, Sanh, Chaumond, Delangue, Moi, Cistac,
  Rault, Louf, Funtowicz, Davison, Shleifer, von Platen, Ma, Jernite, Plu, Xu,
  Scao, Gugger, Drame, Lhoest, and Rush}]{Wolf2019HuggingFacesTS}
Wolf, T.; Debut, L.; Sanh, V.; Chaumond, J.; Delangue, C.; Moi, A.; Cistac, P.;
  Rault, T.; Louf, R.; Funtowicz, M.; Davison, J.; Shleifer, S.; von Platen,
  P.; Ma, C.; Jernite, Y.; Plu, J.; Xu, C.; Scao, T.~L.; Gugger, S.; Drame, M.;
  Lhoest, Q.; and Rush, A.~M. 2019.
\newblock HuggingFace's Transformers: State-of-the-art Natural Language
  Processing.
\newblock \emph{ArXiv} abs/1910.03771.

\bibitem[{Yih et~al.(2011)Yih, Toutanova, Platt, and Meek}]{yih2011learning}
Yih, W.-t.; Toutanova, K.; Platt, J.~C.; and Meek, C. 2011.
\newblock Learning Discriminative Projections for Text Similarity Measures.
\newblock \emph{CoNLL-2011} 247.

\bibitem[{Zhang et~al.(2020)Zhang, Gupta, Tang, Han, Pradeep, Lu, Zhang,
  Nogueira, Cho, Fang et~al.}]{zhang2020covidex}
Zhang, E.; Gupta, N.; Tang, R.; Han, X.; Pradeep, R.; Lu, K.; Zhang, Y.;
  Nogueira, R.; Cho, K.; Fang, H.; et~al. 2020.
\newblock Covidex: Neural Ranking Models and Keyword Search Infrastructure for
  the COVID-19 Open Research Dataset.
\newblock \emph{arXiv preprint arXiv:2007.07846} .

\end{thebibliography}

\section{Appendix}

Here we describe the open-domain datasets we use and also list out the hyperparameter values for all of our experiments.

\subsection{Open-Domain Datasets}

We use SQuAD1.1 to train the synthetic example generator. 
SQuAD2.0 and Natural Questions are used to fine-tune the machine reading comprehension (MRC) model.
Finally, we use Open-NQ to fine-tune the information retrieval (IR) model.

\subsubsection{Natural Questions} NQ \cite{kwiatkowski2019natural} is an English MRC benchmark which contains questions from Google users, and requires systems to read and comprehend entire Wikipedia articles. The dataset contains 307,373 instances in the train set, 7,830 examples in the dev set and 7842 in a blind test set. \citeauthor{lee2019latent}~\shortcite{lee2019latent} create an open version of this dataset, called \textit{Open-NQ}, wherein they only keep questions with short answers and discard the given evidence document. This open version contains 79,168, 8,757 and 3,610 examples in the train, dev and test set, respectively.

\subsubsection{SQuAD} SQuAD1.1 \cite{rajpurkar2016squad} is an extractive MRC dataset containing  questions posed by crowdworkers on a set of Wikipedia articles. All the questions are answerable, with 87,599, 10,570 and 9,533 examples in the train, dev and test set, respectively. SQuAD2.0 \cite{rajpurkar2018know} combines the 100,000+ questions in SQuAD1.1 with over 50,000 unanswerable questions written adversarially by crowdworkers to look similar to answerable ones. It contains 130,319, 11,873 and 8,862 examples in the train, dev and test set, respectively.

\subsection{Hyperparameters}


Tables \ref{tab:hp_syn}, \ref{tab:hp_ir}, \ref{tab:hp_lm}, \ref{tab:hp_mrc} list the hyperparameters for training the example generation, IR, masked language modeling and MRC models, respectively. 

\begin{table}[h]
\small
    \centering
    \begin{tabular}{c|c}
        Hyperparameter & Value \\
        \hline
        Learning rate & 3e-5   \\
        Epochs & 3 \\
        Batch size & 24 \\
        Max source + target sequence length & 1024 \\
        \hline
    \end{tabular}
    \caption{Hyperparameter settings during training the synthetic example generator on SQuAD1.1.}
    \label{tab:hp_syn}
\end{table}

\begin{table}[h]
\small
    \centering
    \begin{tabular}{c|cc}
        Hyperparameter & ICT & Adapted DPR \\
        \hline
        Learning rate & 1e-5 & 1e-5 \\
        Epochs & 6 & 6 \\
        Batch size & 128 & 128 \\
        Warm-up steps & 1237 & 1237 \\
        Max sequence length & 350 & 350 \\
        \hline
    \end{tabular}
    \caption{Hyperparameter settings for the IR experiments in fine-tuning the DPR model with different adaption strategies.}
    \label{tab:hp_ir}
\end{table}

\begin{table}[ht]
\small
    \centering
    \begin{tabular}{c|c}
        Hyperparameter & Value \\
        \hline
        Learning rate & 1.5e-4  \\
        Epochs & 8 \\
        Batch size & 256 \\
        Max sequence length & 512 \\
        Masking rate & 0.15 \\
        \hline
    \end{tabular}
    \caption{Hyperparameter settings during masked language modeling on the CORD-19 collection.}
    \label{tab:hp_lm}
\end{table}

\begin{table}[ht]
\small
    \centering
    \begin{tabular}{c|ccc}
        Hyperparameter & SQuAD2.0 & SynQ & NQ \\
        \hline
        Learning rate & 3e-5  & 1.6e-5 & 1.6e-5 \\
        Epochs  & 2 & 1& 1\\
        Batch size & 8 & 48 & 48\\
        Max sequence length & 384 & 512 & 512 \\
        Max question length & 64  & 18 & 18 \\
        Document stride & 128 & 192 & 192 \\
        \hline
    \end{tabular}
    \caption{Hyperparameter settings during MRC fine-tuning of the language model.}
    \label{tab:hp_mrc}
\end{table}

\end{document}